\documentclass[10pt,letterpaper]{article}
\usepackage[top=0.85in,left=2.75in,footskip=0.75in]{geometry}

\usepackage{amsmath,amssymb}

\usepackage{changepage}

\usepackage{textcomp,marvosym}

\usepackage{cite}

\usepackage{nameref,hyperref}

\usepackage[right]{lineno}

\usepackage[nopatch=eqnum]{microtype}
\DisableLigatures[f]{encoding = *, family = * }

\usepackage[table]{xcolor}

\usepackage{array}

\usepackage{tikz}
\usepackage{pgfplots}
\usepackage{dialogue}
\usetikzlibrary{shapes,shadows,arrows,positioning}
\usepackage{varwidth}
\definecolor{darkgreen}{RGB}{100,180,51}

\newcolumntype{+}{!{\vrule width 2pt}}

\newlength\savedwidth

\raggedright
\usepackage[aboveskip=1pt,labelfont=bf,labelsep=period,justification=raggedright,singlelinecheck=off]{caption}

\makeatletter
\renewcommand{\@biblabel}[1]{\quad#1.}
\makeatother

\usepackage{lastpage,fancyhdr,graphicx}
\usepackage{epstopdf}
\fancyheadoffset[L]{2.25in}
\fancyfootoffset[L]{2.25in}
\begin{document}
\vspace*{0.2in}

\begin{flushleft}
{\Large
\textbf\newline{Performance of ChatGPT-3.5 and GPT-4 on the United States Medical Licensing Examination With and Without Distractions} 
}
\newline
\\
Myriam Safrai\textsuperscript{1,2},
Amos Azaria\textsuperscript{3*}
\\
\bigskip
\textbf{1} Department of Obstetrics, Gynecology and Reproductive Sciences, Magee-Womens
Research Institute, University of Pittsburgh School of Medicine, Pittsburgh, PA 15213, USA

\textbf{2} Department of Obstetrics and Gynecology, Chaim Sheba Medical Center (Tel Hashomer),
Sackler Faculty of Medicine, Tel Aviv University, Tel Aviv, Israel.

\textbf{3} School of Computer Science, Ariel University, Israel
\\
\bigskip




* amos.azaria@ariel.ac.il

\end{flushleft}
\section*{Abstract}
Efforts are being made to improve the time effectiveness of healthcare providers. Artificial intelligence tools can help transcript and summarize physician-patient encounters and produce medical notes and medical recommendations. However, in addition to medical information, discussion between healthcare and patients includes small talk and other information irrelevant to medical concerns. As Large Language Models (LLMs) are predictive models building their response based on the words in the prompts, there is a risk that small talk and irrelevant information may alter the response and the suggestion given. Therefore, this study aims to investigate the impact of medical data mixed with small talk on the accuracy of medical advice provided by ChatGPT. USMLE step 3 questions were used as a model for relevant medical data. We use both multiple choice and open ended questions.
First, we gathered small talk sentences from human participants using the Mechanical Turk platform. Second, both sets of USLME questions were arranged in a pattern where each sentence from the original questions was followed by a small talk sentence. ChatGPT 3.5 and 4 were asked to answer both sets of questions with and without the small talk sentences.
Finally, a board-certified physician analyzed the answers  by ChatGPT and compared them to the formal correct answer. The analysis results demonstrate that the ability of ChatGPT-3.5 to answer correctly was impaired when small talk was added to medical data for multiple-choice questions (72.1\% vs. 68.9\%; p=0.67) and open questions (61.5\% vs. 44.3\%; p=0.01), respectively. In contrast, small talk phrases did not impair ChatGPT-4 ability in both types of questions (83.6\% and 66.2\%, respectively). According to these results, ChatGPT-4 seems more accurate than the earlier 3.5 version, and it appears that small talk does not impair its capability to provide medical recommendations. Our results are an important first step in understanding the potential and limitations of utilizing ChatGPT and other LLMs for physician-patient interactions, which include casual conversations. 



\section*{Introduction}
One of the key, yet most time-consuming, healthcare tasks is charting and creating medical notes, taking daily hours of healthcare providers \cite{oxentenko2010time}. In fact, this task often requires healthcare providers to spend as much, if not more time than they do in direct patient interaction \cite{ammenwerth2009time,fuchtbauer2013emergency}. For example, in a survey, 67\% of the residents reported spending in excess of 4 hours daily on documentation \cite{oxentenko2010time}.Despite the importance of medical notes \cite{harvey2022more,epstein2022goals} no changes have been made to their format, besides having transferred the responsibility of writing them from other medical team members to the physicians \cite{preiksaitis2023chatgpt}. This shift has created a burden for medical providers \cite{apathy2023documentation} and physician burnout \cite{sanderson2020clinical}. Moreover, the recent implementation of electronic health records (EHRs) has significantly increased clinician documentation time \cite{poissant2005impact}, making it the most time-consuming physician activity \cite{hill20134000}. This emphasizes the pressing need to improve the way of charting and making medical notes.

Large Language Models (LLMs) have been suggested as a possible solution, improving healthcare documentation, creating notes, summarizing physician-patient encounters, and even providing meaningful suggestions for further treatments \cite{liu2023utility,cascella2023evaluating}. For example, Chat Generative Pre-trained Transformer 3.5 (ChatGPT-3.5) has been shown to generate a correct diagnosis for 93\% of clinical cases with common chief complaints \cite{hirosawa2023diagnostic} and screening breast cancer with an average correct rate of 88.9\% \cite{rao2023evaluating}. In addition, ChatGTP-3.5 was able to provide general medical information on common retinal disease \cite{potapenko2023artificial}, in almost every subject in genecology \cite{grunebaum2023exciting} and in cancer subjects \cite{johnson2023using}. Moreover, another article demonstrated ChatGPT-3.5's ability to generate clinical letters with high overall accuracy and humanization \cite{ali2023using}. Recent investigations have also shown ChatGPT-3.5's ability to write medical notes \cite{cascella2023evaluating} and to generate a discharge note based on a brief description \cite{patel2023chatgpt}. More recently, a newer version ChatGPT-4 was released. ChatGPT-4 has the ability to process a greater word limit, a stronger ability to solve complex problems, and image recognition \cite{waisberg2023gpt}. This version has additionally shown greater capabilities in terms of clinical evaluation \cite{nori2023capabilities,kasai2023evaluating}. Namely, while ChatGPT-3.5 has obtained a score of 60.9\% on a US sample clinical exam, ChatGPT-4 obtained a score of 89.8\% on the same exam \cite{nori2023capabilities}. A similar result was obtained on the Japanese medical exam, in which ChatGPT-3.5 obtained an average score of 121.3 on the first part of the exam and 149.7 on the second part, while ChatGPT-4 obtained an average score of 167.7 on the first part of the exam and 221.5 on the second part \cite{kasai2023evaluating}.

Following the success of ChatGPT in the medical field, the technology  has been tested to summarize physician-patient encounters \cite{lee2023benefits}. Those appointments between healthcare providers and patients form the foundation of medical care \cite{bickley2012bates}. They necessitate medical evaluations, including the provider's focus on patient needs, obtaining medical histories  \cite{peterson1992contributions,roshan2000study}, conducting physical examinations \cite{woolf2003history,jarvis2023physical} and performing additional tests if necessary \cite{naucler2007human,medlineplus2023}. Moreover, they also entail non-medical tasks such as documenting patient records, organizing notes, and making referrals.

However, since healthcare and patient discussions are unique and based on trust, in addition to medical information, they often include small talk and other information irrelevant to medical concerns \cite{jin2018small,alvaro2022functions}. Those unique exchanges are an important part of the relationship between medical providers and patients and are common among different cultures \cite{wei2023small}. In traditional chinese medicine doctors actively initiate small talks to acquire holistic information for diagnosis and attach great importance to them\cite{wei2023small}. In contrast, such interaction with small talk, has been found to alter the technical skills and performances of medical students, \cite{posner2013too}. These controversies raise concerns regarding their potential impact on LLMs.

As LLMs are predictive models generating their response based on the words in the provided prompt, \cite{bender2021dangers},  there is a risk that small talk and irrelevant information may alter the response and the provided suggestion. Despite the growing number of studies on the potential of using AI for healthcare purposes, to the best of our knowledge, none have assessed this unique aspect of healthcare-patient content interactions and the effect that casual conversation and unrelated information could have on the efficacy of ChatGPT to process medical information and therefore be used to write medical notes summarizing physician-patient interaction. This study aims to investigate the impact of interspersing medical data with casual conversation on the precision of medical recommendations provided by ChatGPT3.5 and ChatGPT-4.

\section*{Material and methods}

\subsection*{Medical Information}
To assess ChatGPT's capabilities in medical reasoning, we evaluate its responses to questions from the United States Medical Licensing Examination (USMLE). This exam has been successfully used to assess the medical logic of LLMs in previous studies \cite{kung2023performance}. Specifically, to evaluate the LLM's proficiency in addressing clinical queries, we selected the Step 3 exam, which is the final examination in the USMLE sequence that qualifies individuals to practice medicine unsupervised. The multiple-choice questions in this exam primarily test knowledge related to diagnosis and clinical management and reflect clinical situations that a general physician might encounter\footnote{https://www.usmle.org/step-exams/step-3/step-3-exam-content}. 

USMLE Step 3 questions were sourced from the dataset provided by Kung et al. \cite{kung2023performance}. Two distinct sets of questions were utilized in the study. The first comprised the original multiple-choice (MC) questions from the USMLE Step 3 exam, while the second presented the same questions in an open-ended (OE) format. Each set contained 122 questions. 

\label{sec:dataset}
\subsection*{Obtaining Small Talk Sentences} 
We conducted a survey on Amazon's Mechanical Turk platform, which allows researchers to recruit participants for various tasks, including online surveys and experiments. Mechanical Turk has gained considerable popularity in recent years as a tool for research due to its efficiency, cost-effectiveness, and the ability to reach a vast pool of participants \cite{paolacci2010running}. 

In our survey, we required the participants to write sentences with at least 10 words to encourage more thoughtful and meaningful responses and reduce the likelihood of individuals providing rushed, brief answers (e.g., "I ate something", "I saw someone", etc.). This is because we aim for participants to produce meaningful sentences that emulate small talk, ensuring they convey information in a casual conversational manner.

The participants were provided the following instructions. ``Please write 5 different sentences as if you were talking to your friend. Each sentence must describe something that has happened to you or an action that you have performed in the past few days. The sentences should not depend on each-other. It is OK to write sentences about simple everyday occurrences (e.g., "1. I sat on a chair on my balcony and looked at the cars passing by."). Each sentence should be at least 10 words long.''

We note that we intentionally framed the small talk in the context of ``talking to a friend" rather than talking to a physician, since we did not want the small talk sentences to have any true influence on the correct answer. By framing the small talk in the context of talking to a friend, we aimed for the correct diagnosis to remain unchanged.

We elicited 35 participants, each provided 5 sentences. This resulted in 175 sentences. The following are some examples of sentences we received from the Mechanical Turk workers:
\begin{enumerate}
    \item I had a great time catching up with my friends at the coffee shop.
    \item I finished reading a great book and I'm looking for my next one.
    \item I biked to the park and watched the birds for an hour.
\end{enumerate}
 All sentences shorter than 10 words were removed. The remaining sentences were converted to a third person's view, to better align with the USMLE format. This resulted in a list of 143 small talk sentences, which are provided in the appendix. 
 
 Converting the three aforementioned sentences to a third person's view, obtains the following:
\begin{enumerate}
    \item The person had a great time catching up with their friends at the coffee shop.
    \item The person finished reading a great book and is looking for their next one.
    \item The person biked to the park and watched the birds for an hour.
\end{enumerate}

\subsection*{Small talk Integration Into Medical Information}
A program was developed that executed the following procedure on the USMLE Step 3 questions. Through sentence tokenization, each question was broken down into individual sentences, and a small talk sentence was inserted. Once processed, each sentence from the USMLE question was followed by a sentence from the small talk file, creating an alternating sequence, as shown in Fig. \ref{fig:question}. 

The final dataset included a total of 488 questions: 122 multiple-choice questions and 122 open-ended questions, each presented with and without small talk.



\begin{figure}[hbpt]
    \centering
\begin{tikzpicture}[
    user/.style={rectangle, draw=blue, fill=blue!20, rounded corners, drop shadow, align=left, font=\small, minimum width=8cm},
    chatgpt/.style={rectangle, draw=red, fill=red!20, rounded corners, drop shadow, align=left, font=\small, minimum width=8cm, minimum height=4cm},
    smalltalk/.style={draw=green, fill=green!20}
]
    \node[user] (user) {
        \begin{varwidth}{10cm}
        A 37-year-old man comes to the emergency department because he has felt nauseated and light-headed for the past hour. \textcolor{darkgreen}{The person is finally getting the hang of a new coding language.} Medical history is significant for esophageal varices secondary to alcohol-related cirrhosis and ascites treated with spironolactone. \textcolor{darkgreen}{The person went for a walk but forgot to take their wallet with them.} He drinks eight to ten alcoholic beverages daily. \textcolor{darkgreen}{The person listened to their mom and tried to understand what she was trying to say.} While you are obtaining additional history, the patient vomits a large volume of bright red blood and becomes difficult to arouse. \textcolor{darkgreen}{The person watched the sunset over the lake.} Vital signs are temperature $36.0^\circ C$ ($96.8^\circ F$), pulse 110/min, respirations 12/min, and blood pressure 90/50 mm Hg. \textcolor{darkgreen}{The person is planning to go to the movies with friends, eager to see a new Marvel movie.} Following initiation of intravenous fluids, what is the most appropriate immediate management?
        \end{varwidth}
    };

\draw[->, thick, blue] (user.west) -- ++(-1,0) node[midway, above, fill=white, xshift=-0.3cm] {Question};
\end{tikzpicture}

\caption{\textbf{Example of a question from the open-ended question dataset with added small talk sentences}. The small talk sentences, added for this illustration, are highlighted in green (the actual dataset does not contain any color highlighting).}
    \label{fig:question}
\end{figure}

\subsection*{ChatGPT Queries}
ChatGPT was prompted using the OpenAI API (in Python). Each question was submitted as a user query without system messages. Each query was submitted separately as a new query, i.e., our program read each question from the file and submitted it to ChatGPT. We used the openai.ChatCompletion.create function with the default parameters\footnote{https://openai.com/blog/openai-api}.

\subsection*{ChatGPT Answers Assessment}
All the responses from ChatGPT to the various datasets were evaluated by a single board-certified physician (MS). For both multiple-choice and open-ended formats, ChatGPT's responses were validated against the official answers of the original multiple-choice questions. 

\subsection*{Statistical Analysis}
Statistical analyses were performed using Python (Scipy 1.10.1). The different group analyses were conducted using the chi2\_contingency function provided by the scipy.stats library. P values less than $0.05$ were considered statistically significant.

\section*{Results}
\begin{figure}[hpbt]
    \centering
    \includegraphics[trim=80 480 80 80, clip, width=0.8\textwidth]{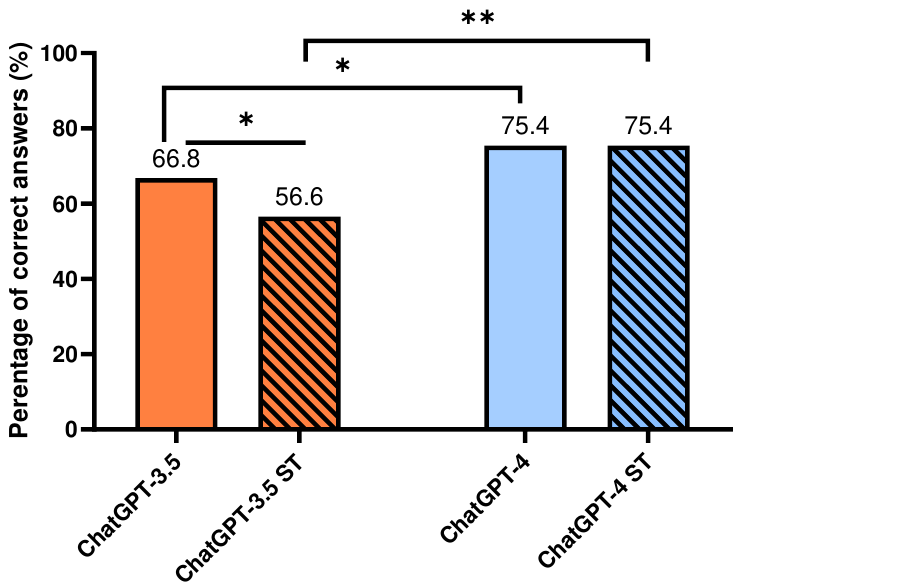}
    \caption{\textbf{Result of the performance of ChatGPT-3.5 and ChatGPT-4 on USMLE datasets with and without small talk.} \emph{Caption:} The figure shows the significant difference in ChatGPT-3.5 and ChatGPT-4 performances' with and without the addition of small talk sentences. In addition, it demonstrates the significant difference in the performance of ChatGPT-3.5 for the datasets with and without small talk addition. ST - Small
talk, with the addition of small talk to the original question. * and ** indicate statistical significance at levels $p<0.05$ and $p<0.001$, respectively.} 
    \label{fig:st_no_st}
\end{figure}

The overall performance of ChatGPT-4 was significantly better than ChatGPT-3.5, with an overall of 61.7\% of correct responses vs. 75.4\% respectively (p. value $< 0.001$). A significantly better score was observed for ChatGPT-4 version when comparing the overall answer to the USMLE question without the addition of small talk (75.4\% vs 66.8\%, p.value $= 0.045$) and (75.4\% vs 56.6\%, p. value $< 0.001$) for the question including small talk addition (Fig. \ref{fig:st_no_st}). In addition, the effect of small talk integration within medical information differs between the two ChatGPT versions. ChatGPT-3.5, showed a clear decrease in the answers' accuracy when small talk sentences were added to the medical data, with a significant decrease of 66.8\% to 56.6\% for all ChatGPT-3.5 answers (p. value $= 0.025$).

\begin{figure}[hpbt]
    \centering
    \includegraphics[trim=80 440 80 100, clip, width=0.8\textwidth]{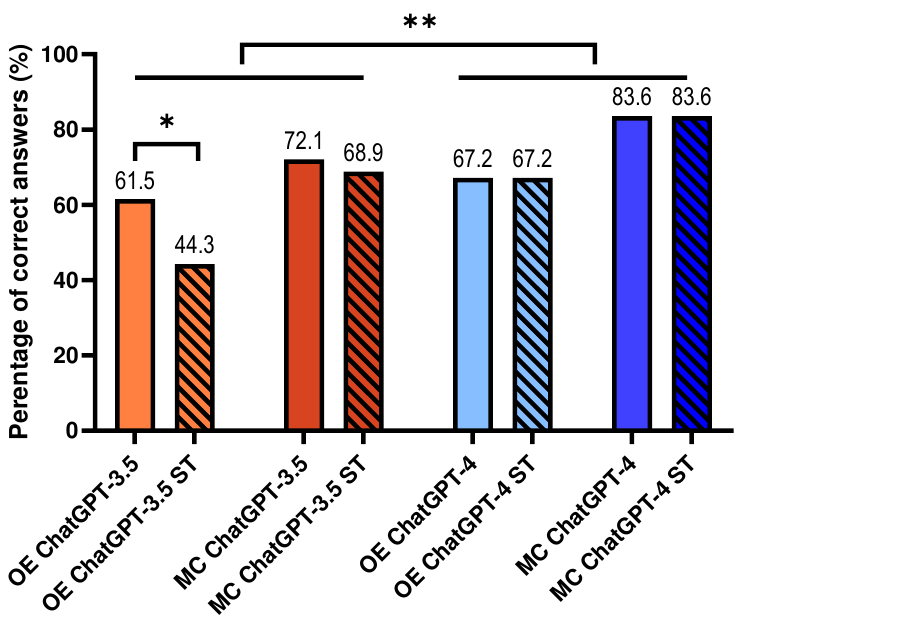}
    \caption{\textbf{Performance of ChatGPT-3.5 and ChatGPT-4 on the two types of USMLE questions, i.e., multiple choice and open ended, with and without small talk.} \emph{Caption:} ChatGPT-4 performed significantly better than ChatGPT-3.5 (p$<0.001$). The small talk seemed to have a larger effect on the performance of ChatGPT-3.5 in the open ended questions. ST - Small
talk, with the addition of small talk to the original question. OE - Open-ended questions, MC - Multiple choice questions.  * and ** indicate statistical significance at levels $p<0.05$ and $p<0.001$, respectively.
}
    \label{fig:oe_mc}
\end{figure}

While looking at each separate data set of questions, the influence of small talk integration on each type of question is more prominent. ChatGPT-3.5 demonstrates a non-significant reduction from 72.1\% to 68.9\% for the multiple-choice questions, while a more considerable and significant drop in performance from 61.5\% to 44.3\% (p. value $=0.01$) was observed for open-ended questions. In contrast, the performance of ChatGPT-4 remained unchanged despite the introduction of small talk, displaying 67.2\% and 83.6\% of correct answers for open and multiple-choice questions, respectively (Fig. \ref{fig:oe_mc}).

Upon closer examination of the answers of ChatGPT to each question, a pattern of error can be observed in ChatGPT responses' when the correct answer is that no further test or investigation was required. For instance, each dataset included two questions, with the correct answer being ``No further evaluation is necessary" or ``No additional study is indicated." Both ChatGPT versions responded incorrectly in the case of open questions suggesting further investigation or treatment regardless in small talk addition. In contrast, when prompted with the dataset of multiple choice of questions, ChatGPT 3.5 answered one of the 2 questions right when no small talk was inserted, and was disturbed by the addition of small talk and responded wrong for both questions after adding small talk. ChatGPT-4 also improved his score on multiple questions and got one correct answer. In contradiction to ChatGPT3.5, its answer was not impaired by adding small talk, and the performance was the same even after adding irrelevant information. 

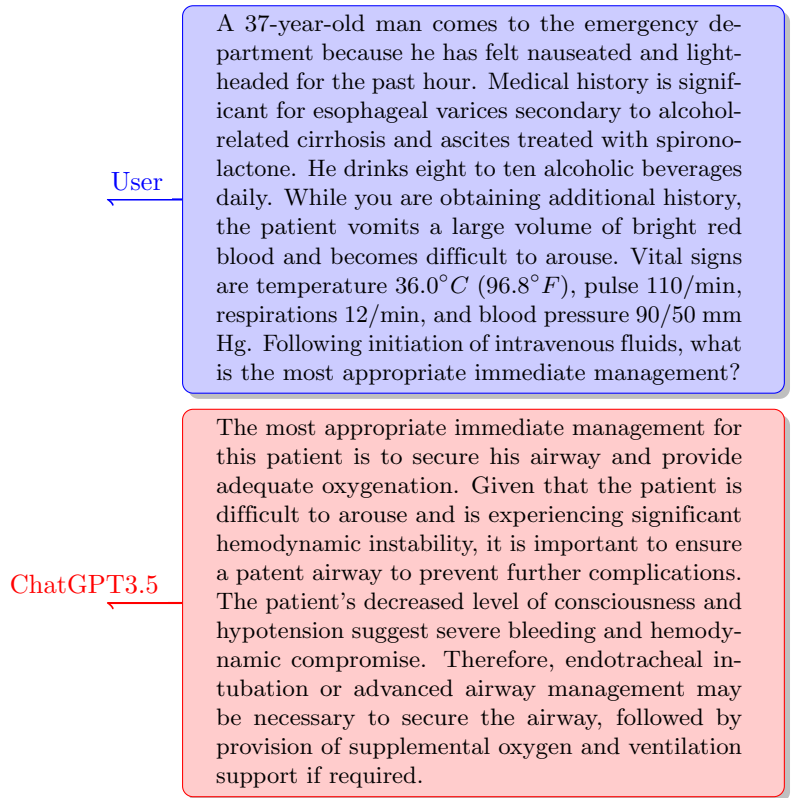
\begin{figure}[ht]
    \centering
\begin{tikzpicture}[
    user/.style={rectangle, draw=blue, fill=blue!20, rounded corners, drop shadow, align=left, font=\small, minimum width=8cm, minimum height=2cm},
    chatgpt/.style={rectangle, draw=red, fill=red!20, rounded corners, drop shadow, align=left, font=\small, minimum width=8cm, minimum height=3.5cm}
]
    \node[user] (user) {
        \begin{varwidth}{7cm}
        A 37-year-old man comes to the emergency department because he has felt nauseated and light-headed for the past hour.  Medical history is significant for esophageal varices secondary to alcohol-related cirrhosis and ascites treated with spironolactone. He drinks eight to ten alcoholic beverages daily.  While you are obtaining additional history, the patient vomits a large volume of bright red blood and becomes difficult to arouse. Vital signs are temperature $36.0^\circ C$ ($96.8^\circ F$), pulse 110/min, respirations 12/min, and blood pressure 90/50 mm Hg.  Following initiation of intravenous fluids, what is the most appropriate immediate management?
        \end{varwidth}
    };

    \node[chatgpt, below=0.2cm of user.south] (chatgpt) {
        \begin{varwidth}{7cm}
        The most appropriate immediate management for this patient is to secure his airway and provide adequate oxygenation. Given that the patient is difficult to arouse and is experiencing significant hemodynamic instability, it is important to ensure a patent airway to prevent further complications. The patient's decreased level of consciousness and hypotension suggest severe bleeding and hemodynamic compromise. Therefore, endotracheal intubation or advanced airway management may be necessary to secure the airway, followed by provision of supplemental oxygen and ventilation support if required.
        \end{varwidth}
    };

\draw[->, thick, blue] (user.west) -- ++(-1,0) node[midway, above, fill=white, xshift=-0.1cm] {User};
\draw[->, thick, red] (chatgpt.west) -- ++(-1,0) node[midway, above, fill=white, xshift=-0.8cm] {ChatGPT3.5};
\end{tikzpicture}
    \caption{\textbf{ChatGPT-3.5 provides a correct answer to a question that does not include small talk interference.}}
    \label{fig:correct}
\end{figure}

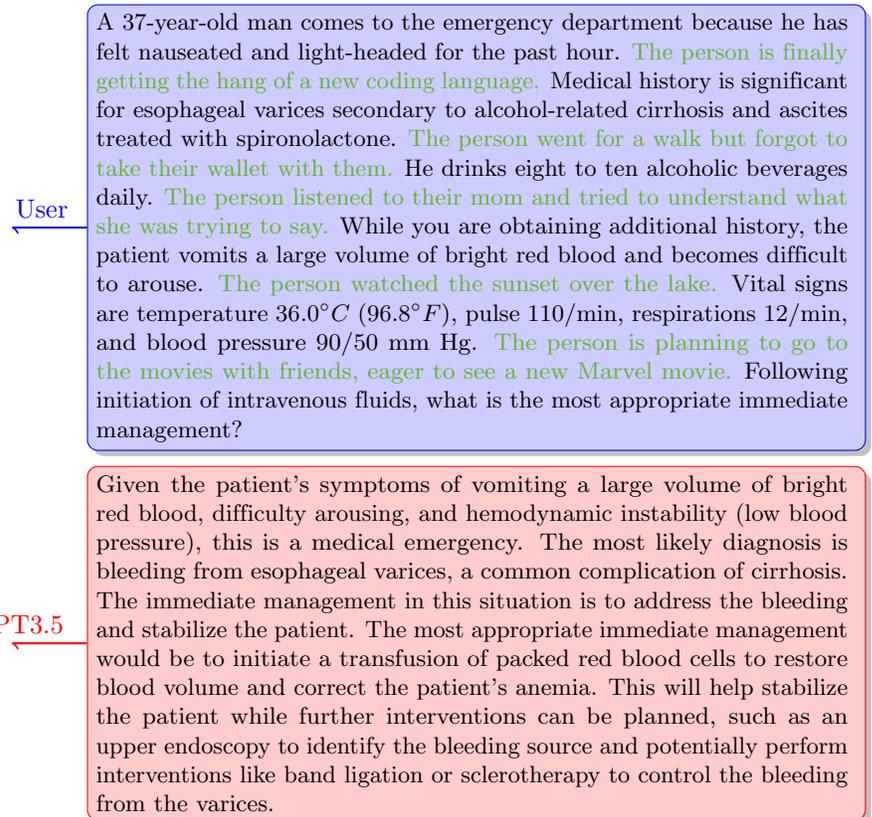
\begin{figure}[hbpt]
    \centering
\begin{tikzpicture}[
    user/.style={rectangle, draw=blue, fill=blue!20, rounded corners, drop shadow, align=left, font=\small, minimum width=1cm},
    chatgpt/.style={rectangle, draw=red, fill=red!20, rounded corners, drop shadow, align=left, font=\small, minimum width=8cm, minimum height=4cm},
    smalltalk/.style={draw=green, fill=green!20}
]
    \node[user] (user) {
        \begin{varwidth}{10cm}
        A 37-year-old man comes to the emergency department because he has felt nauseated and light-headed for the past hour. \textcolor{darkgreen}{The person is finally getting the hang of a new coding language.} Medical history is significant for esophageal varices secondary to alcohol-related cirrhosis and ascites treated with spironolactone. \textcolor{darkgreen}{The person went for a walk but forgot to take their wallet with them.} He drinks eight to ten alcoholic beverages daily. \textcolor{darkgreen}{The person listened to their mom and tried to understand what she was trying to say.} While you are obtaining additional history, the patient vomits a large volume of bright red blood and becomes difficult to arouse. \textcolor{darkgreen}{The person watched the sunset over the lake.} Vital signs are temperature $36.0^\circ C$ ($96.8^\circ F$), pulse 110/min, respirations 12/min, and blood pressure 90/50 mm Hg. \textcolor{darkgreen}{The person is planning to go to the movies with friends, eager to see a new Marvel movie.} Following initiation of intravenous fluids, what is the most appropriate immediate management?
        \end{varwidth}
    };

    \node[chatgpt, below=0.2cm of user.south] (chatgpt) {
        \begin{varwidth}{10cm}
        Given the patient's symptoms of vomiting a large volume of bright red blood, difficulty arousing, and hemodynamic instability (low blood pressure), this is a medical emergency. The most likely diagnosis is bleeding from esophageal varices, a common complication of cirrhosis. The immediate management in this situation is to address the bleeding and stabilize the patient. The most appropriate immediate management would be to initiate a transfusion of packed red blood cells to restore blood volume and correct the patient's anemia. This will help stabilize the patient while further interventions can be planned, such as an upper endoscopy to identify the bleeding source and potentially perform interventions like band ligation or sclerotherapy to control the bleeding from the varices.
        \end{varwidth}
    };

\draw[->, thick, blue] (user.west) -- ++(-1,0) node[midway, above, fill=white, xshift=-0.1cm] {User};
\draw[->, thick, red] (chatgpt.west) -- ++(-1,0) node[midway, above, fill=white, xshift=-0.8cm] {ChatGPT3.5};
\end{tikzpicture}
    \caption{ChatGPT-3.5 incorrectly responds to a question mixed with small talk (highlighted in green).}
    \label{fig:smalltalk-wrong}
\end{figure}

In other questions, where the correct answer was a diagnosis or treatment, the addition of small talk impaired ChatGPT-3.5 performance. For example, as seen in Fig. \ref{fig:correct}, the response was correct before adding small talk. However, as shown in Fig. \ref{fig:smalltalk-wrong}, once small talk phrases were added to the question, ChatGPT-3.5 failed and provided an incorrect response. Interestingly, even though the small talk caused ChatGPT to respond incorrectly, it does not explicitly mention any of the small talk information in its answer and does not explain its wrong answer based on the specific interference added to this question.

\section*{Discussion}

The primary purpose of this study was to investigate the effect of the addition of small talk to medical data on the accuracy of medical advice provided by ChatGPT. First, as expected, ChatGPT-4 outperforms ChatGPT-3.5 with an overall higher score for open and multiple-choice questions. This matches the expectation as ChatGPT-4 is a more advanced version and has been shown to outpass ChatGPT-3.5 on multiple-choice questions in the US and Japan medical exams \cite{nori2023capabilities,kung2023performance}. However, this is the first study to show a similar improvement in the capacity of ChaGPT-4 to surpass ChatGTP-3.5, giving medical recommendations to open questions that simulate daily clinical needs. The high score of almost three-quarters of correct answers of ChatGPT-4 for open questions in our study indicates its ability to process medical information. These findings suggest the capacity of ChatGPT-4 to respond and provide medical advice and demonstrate its potential future use in the medical field. 

When evaluating the effect of small talk addition to the different datasets, ChatGPT-3.5 showed a slight drop in performance for multiple-choice questions and a significant one in answering the open-ended questions following the addition of small talk. In contrast, ChatGPT-4's performance was consistent regardless of small talk, with stable accuracy rates for both question types. To our knowledge, this is the first study evaluating the effect of small talk on ChatGPT and other LLMs' efficacy in processing medical information in the context of unformal or irrelevant information. Our study demonstrates the various impacts of adding small talk on different versions of ChatGPT. It implies that the addition of small talk does not impair ChatGPT-4 performance in processing medical data, which can provide the same accuracy in medical recommendations as in `medical only' conversation. During a provider-patient interaction, irrelevant information is often mixed with medical data, which needs to be processed and summarized in contrast to the small talk. It has been demonstrated in a previous study that ChatGPT can summarize and provide a note for 'medical only' physician–patient encounters \cite{lee2023benefits}. Therefore, our data suggests that ChatGPT-4 can assist in this task without being impaired by a patient-provider casual discussion that might occur and be provided to ChatGPT in a transcript. These findings provide important answers for medical practitioners and LLM developers regarding the potential of the implication of ChatGPT and other LLMs as a tool in medicine. This is especially important as it is predicted that chatbots will be used by medical professionals, as well as by patients, with increasing frequency \cite{lee2023benefits}.

The analysis of the exact scoring of ChatGPT in our study demonstrates that ChatGPT-3.5 answered 72.1\% of the multiple-choice questions correctly without small talk integration. This score is higher than the one reported by Kung et al. \cite{kung2023performance} ranging from 68.8\% to 61.5\%. 
It should be noted, however, that our study was conducted approximately $8$ months after the original assessment. A possible explanation for this difference is that ChatGPT, as an Artificial Intelligence system, has learned and adapted from the data. As it encounters more information, it refines its models, which often leads to improved performance and accuracy \cite{chen2023chatgpt}. It is plausible that the elevated scores observed in our research can be attributed to a marked learning enhancement. These findings likely underscore the continuous improvement of ChatGPT over time. We are optimistic that subsequent studies will yield even more favorable outcomes, enhancing ChatGPT's ability to offer even better medical recommendations and furnish dependable support to healthcare providers in medical record documentation.

Each dataset included 2 questions, where the correct answer was that no further investigation was required. Both versions of ChatGPT answers to the open-ended questions were wrong. In contrast, for multiple choice questions, ChatGPT3.5 had one of the two questions answered correctly if no small talk was added and both were wrong after this addition, whereas ChatGPT-4 was not influenced by small talk addition constantly answers to one of two questions correctly. Our study is the first to report the need and complexity of LLMs to respond to those types of questions. Those types of answers  are crucial in medicine as patients can be easily referred to countless further tests and investigations, burdening the patients and the medical system \cite{carpenter2015overtesting}. Those queries challenge LLMs for whom specific wording of the prompt influence dramatically the answer provided \cite{qin2021learning}. In those examples, asking what should be the next step may induce that a next step is indeed required. That finding demonstrates the complexity of using ChatGPT in different queries and the need to acknowledge the limit of this technology at this current development.

Finally, we seek to analyze the cause of the small talk disturbance to ChatGPT-3.5 processing. We hypothesized that adding different subjects and specific words would engender a failure in the process of ChatGPT-3.5. However, while the presence of small talk impaired the performance of ChatGPT-3.5 for the datasets of questions,  the answer provided by ChatGPT-3.5 did not explain the wrong answer based on a specific subject or word included in the small talk. This result is concerning, as by delivering incorrect responses but still not mentioning any unrelated information, it may be difficult for a health provider reviewing the answers to pinpoint errors.

Our study has several limitations. The prominent one is that it is challenging to mimic the small talk that is occurring between the health provider and a patient. In our model, we framed the small talk in the context of ``talking to a friend" rather than a physician to avoid bias and integration of medical terms. However, in practice, the patient will be talking to a physician; thus, even the small talk may resemble medical information being conveyed. Such small talk might deteriorate the performance of ChatGPT3.5 and might even affect the performance of ChatGPT-4, which, in our analysis, seemed immune to small talk.  

In addition, in our work, the small talk sentences and the medical information were added in an alternating sequence to USMLE questions, each small talk sentence was added as a standalone piece of information. However, in medical practice, the transcript of physician-patient interaction may be much longer than a USMLE question, and the small talk might be structured differently. USMLE question has been used previously to assess medical data processing \cite{kung2023performance}, reinforcing the use of a dataset for such a purpose and allowing us to compare our results. Nevertheless, it is possible that different patterns of small talk integration on different scripts might have various effects on ChatGPT's ability to provide medical counsel. We would also like to stress that this work focuses on both medical information and small talk conveyed in text; however, in practice, the irrelevant information can be conveyed in different modes, such as images (either medical-related images or pictures of the patient's family, pets, etc.) or sounds (either caused by a medical condition of the patient, or the patient laughing as a response to a joke, imitating their boss, etc.). Despite this, the present analysis provides important new information about the impact of the most common way of communicating\cite{patel2023chatgpt,roshan2000study}, including irrelevant information, in physician–patient encounters on the ability of the different versions of ChatGPT to provide medical advice.

Another potential limitation of this study is that it focuses on ChatGPT-only and has not assessed different LLMs and therefore cannot be generalized to other forms of LLMs. Future research could thus attempt to investigate whether the addition of small talk interferes with other LLMs (such as BERT, Cloude, LLAMA-1, and LLAMA-2)  ability to provide medical advice 

In this paper, we took the first step toward understanding the performance of the two ChatGPT versions, when faced with physician-patient interactions including medical mixed with irrelevant information. Those unique interactions raised a challenge to discern the impact of casual conversations on the accuracy and reliability of medical recommendations made by these LLMs. This analysis shows that while ChatGPT-3.5 performance was significantly impaired by small talk addition, ChatGPT-4 performance was not affected. The results have significant implications for the integration of LLMs into medical practice. In addition, LLM developers, and especially healthcare providers must be aware of these limitations. It should be emphasised that while LLMs can assist in many tasks, it is crucial to critically review and evaluate the suggestions and notes generated, particularly in the context of patient interactions filled with non-medical content.






\bibliography{references}

\end{document}